%% file: main.tex
\documentclass[11pt,a4paper]{wan}
\usepackage{amsmath,amssymb,amsfonts,bm}
\usepackage{graphicx,booktabs,tabularx,array}
\usepackage{enumitem,microtype,placeins,needspace}
\usepackage{colortbl}
\definecolor{TableGroup}{rgb}{0.95,0.95,1.00}
\definecolor{TableBest}{rgb}{0.925,0.700,0.775}
\definecolor{TableSecond}{rgb}{0.9625,0.8500,0.8875}
\newcommand{\bestcell}[1]{\cellcolor{TableBest}\textbf{#1}}
\newcommand{\secondcell}[1]{\cellcolor{TableSecond}#1}
\newcommand{\ranklegend}{\colorbox{TableBest}{best} and \colorbox{TableSecond}{second-best}}
\newsavebox{\rsigenRefreshTableBox}
\newsavebox{\rsigenRecursionCaptionBox}
\newsavebox{\rsigenRefreshCaptionBox}
\newlength{\rsigenPanelBodyHeight}
\newlength{\rsigenPanelCaptionHeight}
\usepackage{hyperref}
\hypersetup{
  pdftitle={RSIgen: Reciprocal Self-Improvement for Agentic Visual Generation},
  pdfauthor={Yidi Liu, Haozhe Wang, Jie Xiao, Kai Zhu, Xueyang Fu, Zheng-Jun Zha},
  pdfsubject={RSIgen arXiv paper (Wan template)}}
\AddToHook{env/figure/begin}{\setlength{\abovecaptionskip}{6pt}\setlength{\belowcaptionskip}{0pt}}
\AddToHook{env/table/begin}{\setlength{\abovecaptionskip}{2pt}\setlength{\belowcaptionskip}{4pt}}

\input{math_commands}
\title{RSIgen: Reciprocal Self-Improvement\\
for Agentic Visual Generation}
\author[1]{Yidi Liu}
\author[3,*]{Haozhe Wang}
\author[2]{Jie Xiao}
\author[2]{Kai Zhu}
\author[1,*]{Xueyang Fu}
\author[1]{Zheng-Jun Zha}
\affiliation[1]{University of Science and Technology of China\\}
\affiliation[2]{Wan Team, Alibaba Group\\}
\affiliation[3]{The Hong Kong University of Science and Technology}
\contribution[*]{Corresponding authors}
\date{October 12, 2026}
\abstract{\input{sections/00_abstract}}

\fancypagestyle{firstheader}{%
  \fancyhf{}
  \fancyhead[L]{\small\sffamily\color{wanMuted}RSIgen}
  \fancyhead[R]{\includegraphics[height=0.64cm]{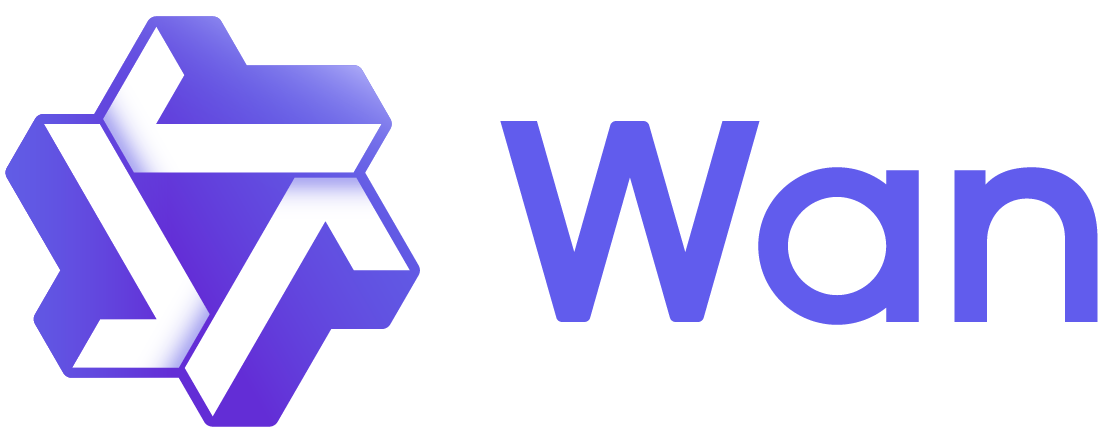}}
  \fancyfoot[C]{\small\thepage}
  }

\begin{document}
\maketitle
\thispagestyle{firstheader}

\raggedbottom
\input{sections/01_introduction}
\input{sections/02_related_work}
\input{sections/03_method}
\input{sections/04_experiments}
\input{sections/05_results}

\FloatBarrier
\input{sections/06_discussion}
\input{sections/07_conclusion}
\FloatBarrier
\label{main-body-end}
\setlength{\bibsep}{3pt plus 1pt}
\bibliographystyle{assets/plainnat}
\bibliography{references}
\end{document}

%% file: math_commands.tex
\newcommand{\Pool}{\mathcal{D}_{\mathrm{pool}}}
\newcommand{\Acc}{\mathcal{D}_{\mathrm{acc}}}
\newcommand{\Seal}{\mathcal{D}_{\mathrm{seal}}}

\setlist[itemize]{leftmargin=*,itemsep=1pt,topsep=3pt}
\setlist[enumerate]{leftmargin=*,itemsep=1pt,topsep=3pt}

%% file: sections/00_abstract.tex
Experience from agentic visual generation can improve the native understanding (agent) and generation roles of a unified multimodal backbone, but task success alone does not determine who should learn. Generator updates can also make action supervision stale. We introduce RSIgen, a task-centric reciprocal self-improvement framework linking learning allocation, supervision refresh, and recursive curriculum construction. Dynamic Responsibility Routing (DRR) uses matched interventions on evidence, inputs, and supported actions to select recipients and construct compatible training targets. After generator learning, RSIgen refreshes agent supervision by comparing tested actions' relative utility with branch inputs fixed; retained roles construct the next curriculum. Across BAGEL, Qwen-Image, and Wan-Image-7B, RSIgen improves AgentGen-Bench Full Overall9 by 3.2--3.3 points over a three-cycle SearchGen adaptation with action costs. Wan controls support allocation gains at matched role-level training exposure, refresh gains at a matched adaptation budget, and benefits from updated curriculum sources while recipients continue learning. Blind evaluation on 200 real-user requests yields pooled tie-adjusted preferences of 70.0\% for task completion and 64.9\% for visual quality against the same baseline.

%% file: sections/01_introduction.tex
\section{Introduction}
\label{sec:introduction}
Agentic visual generation turns an image request into decisions about evidence acquisition, input construction, generation, and repair~\citep{unifyagent,qwenagent}. We study the \emph{native-role setting}: the agent and generator are the understanding and generation roles of the same unified multimodal backbone. Task success depends on both the agent's orchestration and the generator's ability to realize the resulting inputs.

A useful action and a useful training target answer different questions. Consider an architecture poster that places buildings in the wrong chronological order. If correcting the layout instruction fixes the image while the evidence and generator stay unchanged, the correction can train the agent. A validated search-assisted image of the same buildings, paired with an evidence-free request, may instead teach stable historical knowledge to the generator. The final image alone does not identify the learning recipient or supply a compatible target.

\begin{figure}[!t]
\centering
\includegraphics[width=1\linewidth]{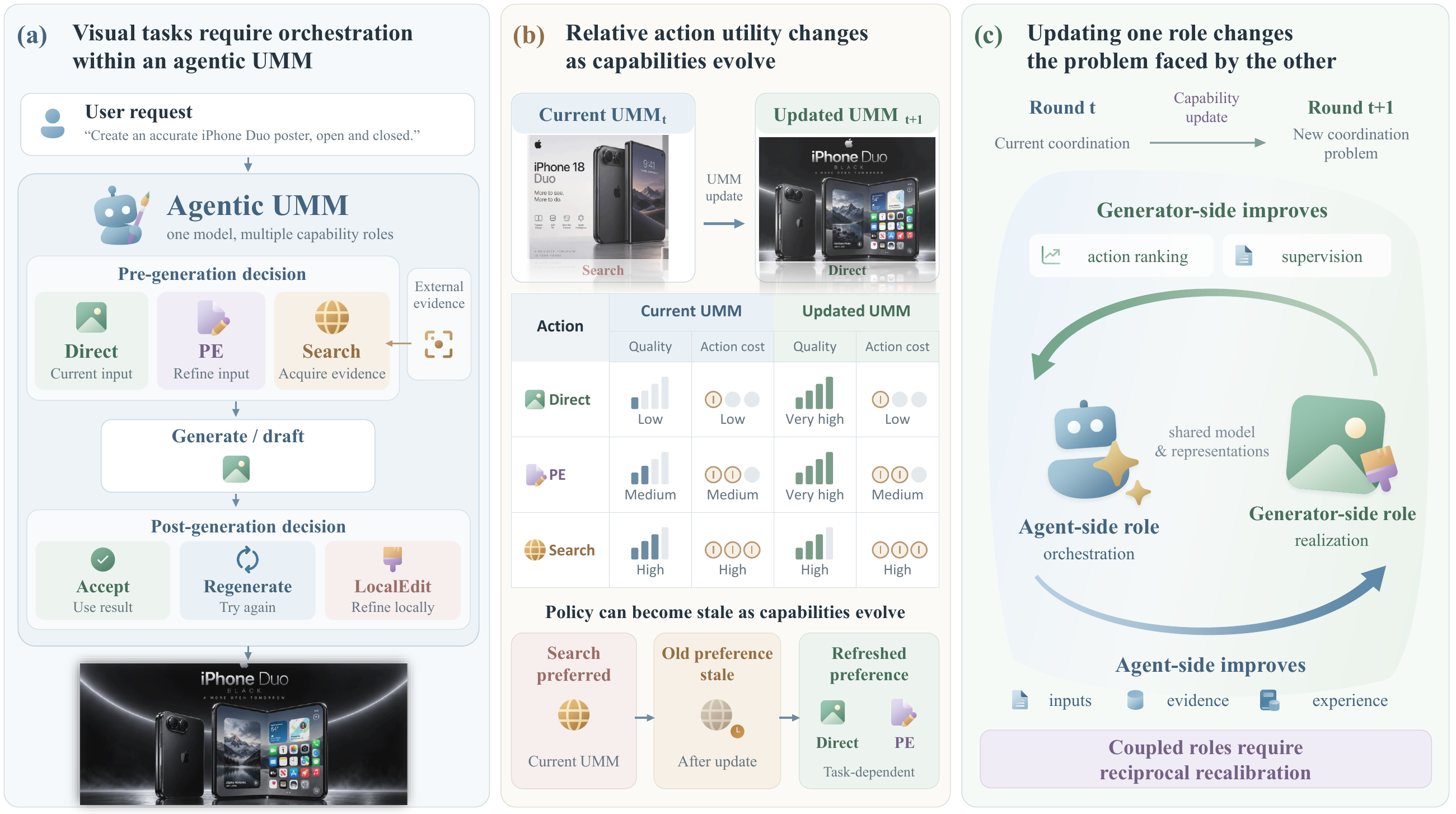}
\caption{Reciprocal learning in agentic visual generation. (a) Agent orchestration and generator realization are roles within a unified multimodal model (UMM); PE denotes prompt enhancement. (b) Generator learning changes action utility. (c) Agent learning changes inputs and evidence for later curricula. Images and quality levels are illustrative.}
\label{fig:overview}
\end{figure}

Learning by either role reshapes its partner's next learning problem. If Direct succeeds on reserved uses of newly learned knowledge while Search remains reliable, an earlier search-required label may become stale. Shared architecture does not automatically recalibrate the agent's policy to the generator's updated capabilities. This \emph{capability--policy calibration} aligns the agent's evidence use with the generator's evolving capabilities. Better agent inputs and evidence, in turn, change the generator's next learning opportunities (Figure~\ref{fig:overview}).

SearchGen already teaches the generator and then recalibrates its reasoner~\citep{searchgen}, while GenEvolve distills contrasting trajectory experience into an evolving agent~\citep{genevolve}. Building on these advances, we study three linked decisions: \emph{Who should learn? Which actions remain useful after an update? What should the updated roles learn next?}

We introduce \textbf{RSIgen}, a framework for reciprocal self-improvement. Dynamic Responsibility Routing (DRR) uses matched interventions on evidence, inputs, and supported actions to select learning recipients and skill subtypes and construct compatible targets.

\Needspace{5\baselineskip}
After generator learning, RSIgen repeats affected action comparisons with the agent, branch inputs, evidence, evaluator, and image seeds fixed. These comparisons refresh supervision according to updated relative action utility across acquisition, input construction, stopping, and supported repair. The agent then learns with the updated generator fixed, and the retained roles construct the next curriculum.

Across BAGEL, Qwen-Image, and Wan-Image-7B, RSIgen improves AgentGen-Bench Full Overall9 by 3.2--3.3 points over SearchGen-3 + costs, our three-cycle SearchGen adaptation with action costs. Blind evaluation on 200 real-user requests yields pooled tie-adjusted preferences of 70.0\% for task completion and 64.9\% for visual quality against this baseline.

Our contributions are:
\begin{itemize}[leftmargin=*,nosep]
\item \textbf{Diagnosis-guided allocation:} distinguish execution choices from learning-recipient allocation and construct compatible targets, with gains at matched role-level training exposure.
\item \textbf{Supervision refresh:} re-evaluate tested pre- and post-generation actions after generator learning, improving adaptation over current-generator on-policy learning at a shared total budget.
\item \textbf{Reciprocal curriculum analysis:} establish the benefit of updated curriculum sources through controls that freeze those sources while keeping recipient learning active.
\end{itemize}

%% file: sections/02_related_work.tex
\section{Related Work}
\label{sec:related-work}
\paragraph{Agentic visual generation.}
Retrieval-augmented generation addresses knowledge and reference requirements through external information. Re-Imagen conditions synthesis on retrieved image--text pairs, while retrieval-augmented diffusion uses retrieved image embeddings~\citep{reimagen,rdm}. Agentic systems decide how to acquire and use this information: Gen-Searcher learns search decisions with a frozen generator and image/text feedback; ToolArtist learns native multimodal tool use; Qwen-Image-Agent organizes planning, search, and memory; and Unify-Agent jointly supervises textual trajectories and image generation~\citep{gensearcher,toolartist,qwenagent,unifyagent}. RSIgen studies how experience from this process should train different capabilities and how action supervision should change when the generator also learns.

\paragraph{Experience reuse and generator-dependent adaptation.}
SearchGen trains the generator with Diffusion-DPO and recalibrates the reasoner through positive-advantage rejection finetuning with the updated generator~\citep{searchgen}. RSIgen adds matched diagnosis for recipient-specific targets and explicit re-evaluation of acquisition and repair supervision. Our SearchGen-3 + costs baseline repeats this DPO--RFT sequence for three cycles with trajectory costs. GenEvolve extracts structured experience from contrasting same-request trajectories and supplies it to a privileged teacher for token-level distillation~\citep{genevolve}. RSIgen allocates validated experience across learning roles and refreshes their action supervision as capabilities evolve.

\paragraph{Reciprocal multimodal self-improvement.}
HermesFlow aligns cross-task preferences; UniRL and UniCorn construct multimodal learning tasks from generated experience; SUDER introduces dual self-rewards; and Ask, Solve, Generate alternates role-specific updates~\citep{hermesflow,unirl,unicorn,suder,asg}. Internal Gap revisits previously underutilized samples as capabilities improve, while Dreaming in Flow combines flow-level feedback with visual replay grounding~\citep{internalgap,dreamingflow}. RSIgen couples the roles through curriculum construction and action supervision: generator learning changes useful agent decisions, and agent learning changes the experience available for later generator training. Its construction-source controls evaluate the latter direction while recipients continue learning.

\paragraph{Optimization and feedback.}
Flow matching, DPO, Diffusion-DPO, and GRPO provide the underlying optimization tools~\citep{flowmatching,dpo,diffusiondpo,deepseekmath}. Flow-GRPO, T2I-R1, UniGRPO, and UniRL-Zero apply outcome-driven optimization to generation or multimodal reasoning~\citep{flowgrpo,t2ir1,unigrpo,unirlzero}. Understanding-based rewards provide feedback for multimodal learning~\citep{gvu,unifiedreward,srum}. RSIgen uses completed-image rewards and trajectory costs for RL, with probes selecting curricula and validated trajectories providing action supervision. COMA assigns credit within policy optimization, whereas DRR assigns compatible learning targets before optimization~\citep{coma}. SPIN and self-rewarding models also use model-generated supervision~\citep{spin,selfreward}. Agent self-improvement also includes changes to implementations and improvement mechanisms~\citep{dgm,hyperagents}. RSIgen focuses on reciprocal updates to understanding and generation through a shared training protocol.

\paragraph{Learning and coordination.}
SRUM freezes its understanding evaluator while training generation, and Endogenous Reprompting freezes generation while learning evaluation and reprompting~\citep{srum,seer}. GEMS and Mind-Brush broaden agentic generation through memory, skills, search, and reasoning~\citep{gems,mindbrush}. DAgger addresses the distribution induced by a changing policy~\citep{dagger}; RSIgen additionally considers changing action utility induced by generator learning.

\FloatBarrier

%% file: sections/03_method.tex
\section{Task-Centric Reciprocal Self-Improvement}
\label{sec:method}
\begin{figure}[!t]
\centering
\includegraphics[width=1\linewidth]{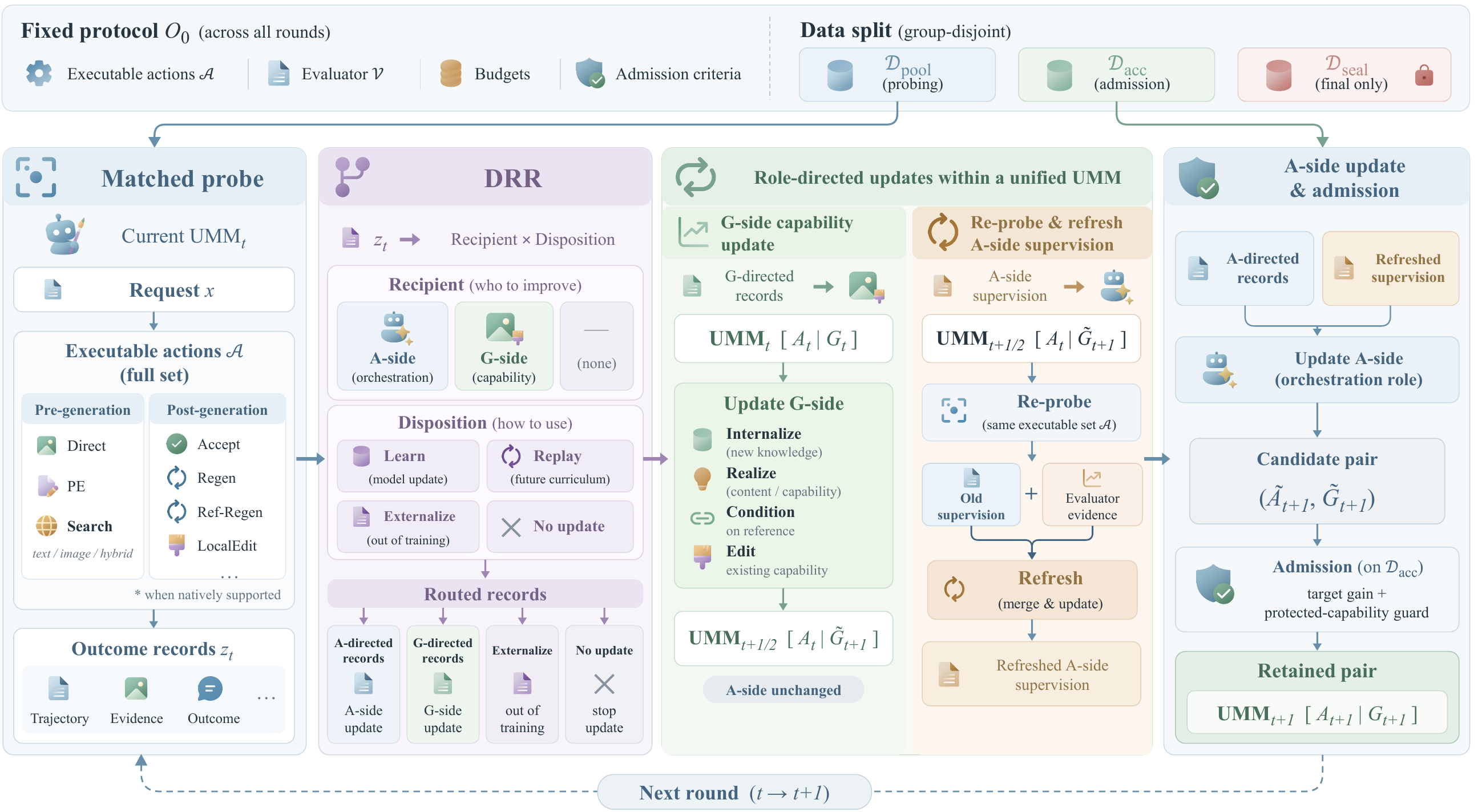}
\caption{RSIgen's learning procedure. DRR compiles matched probes into typed curricula. Generator learning precedes re-probing and agent learning; retained roles construct the next curriculum. Pool, admission, and sealed evaluation data have separate roles. UMM denotes the unified model's understanding/orchestration and generation roles. Evaluator feedback and capability memory support training.}
\label{fig:framework}
\end{figure}

RSIgen links learning-recipient allocation, supervision refresh, and recursive curriculum construction through a shared training protocol (Figure~\ref{fig:framework}).

\subsection{Task, capability roles, and outcome rewards}
At round $t$, $S_t=(A_t,G_t,M_t)$ comprises the native understanding/agent and generation roles of a unified multimodal backbone, plus outer-loop capability memory. Each update changes its recipient while holding the other role's effective function fixed. Parameters follow the backbone; Wan uses full-parameter updates to the trained role. For request $x$,
\begin{equation}
x\xrightarrow{A_t:\mathrm{Think}}\text{decision}
\xrightarrow{A_t:\mathrm{Action}}(p,e)
\xrightarrow{G_t}I\xrightarrow{\mathcal E_0}y.
\label{eq:chain}
\end{equation}
The agent supplies prompt $p$ and optional evidence/reference $e$; $\mathcal E_0$ scores the completed image. The agent observes images and tool results. Evaluator scores and $M_t$ guide training.

Pre-generation actions are Direct, PromptEnhance, TextSearch, ImageSearch, and HybridSearch. Post-generation actions comprise accept, regenerate, reference-conditioned regenerate, and available LocalEdit operations.

Both RL stages use completed-image outcomes. With request-applicable checks $\mathcal J(x)$ and normalized weights $w_j$,
\begin{align}
R_G(I,x,e)&=\sum_{j\in\mathcal J(x)}w_jv_j(I,x,e)-\lambda_{\mathrm{viol}}c_{\mathrm{viol}}(I,x,e),\label{eq:rewardg}\\
R_A(\tau,x)&=R_G(I_{\mathrm{final}},x,e_{\mathrm{final}})
-\lambda_s C_s(\tau)-\lambda_g C_g(\tau)-\lambda_l C_l(\tau).\label{eq:rewarda}
\end{align}
Checks $v_j$ evaluate image requirements; $c_{\mathrm{viol}}$ penalizes predefined violations. Factual/reference targets are anchored independently of agent-selected evidence. $C_s$ counts searches, $C_g$ generations/edits, and $C_l$ horizon-normalized decisions. Probe scores construct curricula; reference/KL regularization belongs to optimization.

The objective is held-out quality $Q(A,G)$ under training/inference budgets and a protected-capability guard. Protocol $\Omega_0$ fixes actions, compilation, update stages, rewards, and admission. Pool $\Pool$, admission $\Acc$, and sealed evaluation $\Seal$ remain source/request/artifact-group disjoint, including derived curricula and replay. Matched probes guide learning-recipient allocation.

\subsection{Who should learn: diagnosis and curriculum compilation}
\input{sections/13_conceptual_diagnosis}
Budgeted probes cover anchors, changed capability regions, uncertain comparisons, and under-covered requests. Matched interventions compare Direct, prompt enhancement, search modalities, Oracle evidence, and supported repairs:
\begin{equation}
q_t(x,a)=\frac{1}{K}\sum_{k=1}^K s(I_t^{(k)}(x,a),x),
\label{eq:probe}
\end{equation}
where $s$ scores the completed image. Comparisons fix the request, peer versions, evaluator, unaffected inputs, and image seeds. Each branch's retrieval artifacts are frozen for replay. Empty searches retain their outcomes and costs.

Input checks validate request fidelity, evidence coverage, entity/reference binding, and interface compatibility. Independently validated Oracle evidence diagnoses evidence adequacy. Failed checks trigger reconstruction or no-update.

Dynamic Responsibility Routing (DRR) maps the versioned probe record $z_t$ to a recipient, skill subtype, and disposition: learn, replay, externalize, or no-update (Figure~\ref{fig:diagnosis}; Table~\ref{tab:routing-rules}). Oracle repair of inadequate retrieval suggests agent acquisition; verified prompt corrections suggest agent construction. Residual failure under valid, requirement-complete inputs suggests generator realization.

Successful search also offers a learning use: Direct failure on stable reusable knowledge with validated Search success and no material Oracle gain suggests generator internalization. Retrieval supplies volatile facts and instance-specific references. Refresh tests whether learning supports no-search supervision.

For opportunity $o=(x,z_t)$, let $d_{o,r}$ be a compatible input--target pair for recipient/subtype $r\in\mathcal R(o)$. DRR selects $\rho(o)\in\mathcal R(o)$ or no-update and compiles $\mathcal C_r=\{d_{o,r}:\rho(o)=r\}$. Generator SFT requires a validated admissible image or successful generation target. Pool expansion independently validates each added example's inputs, subtype, and target.

In the poster example, agent construction pairs the request, evidence, and original input with the executed correction. Internalization pairs an evidence-free student request with a validated assisted image:
\begin{equation}
(p^{\mathrm{ctx}},e^*)\longrightarrow I^*,
\qquad G:(p^{\mathrm D},\varnothing)\longrightarrow I^*.
\label{eq:distill}
\end{equation}
The student preserves the request and entity identity but excludes retrieved facts and descriptions. Reference-following records retain their conditions. Generator curricula cover internalization, realization, reference following, and supported editing. Compatible flow generators use SFT/conditional flow matching and then Flow-GRPO for open-ended verifiable skills; internalization uses SFT only.

\paragraph{Routing precedence and admissible targets.}
DRR resolves upstream input failures before considering downstream realization. Regressions replay validated targets for their original recipient and subtype. Table~\ref{tab:routing-rules} summarizes the resulting routes.

\begin{table}[!htbp]
\centering\footnotesize
\caption{From probe outcomes to typed curricula. Each route pairs validated inputs with a recipient-compatible target. Oracle supports diagnosis; agent targets use executed trajectories.}
\label{tab:routing-rules}
\begin{tabularx}{\linewidth}{@{}>{\raggedright\arraybackslash}p{0.36\linewidth}>{\raggedright\arraybackslash}p{0.22\linewidth}>{\raggedright\arraybackslash}X@{}}
\toprule
Probe pattern & Recipient/subtype & Compiled training object\\
\midrule
Oracle repairs inadequate retrieval & Agent: acquisition & Valid query and evidence-selection trajectory\\
Corrected input repairs the output & Agent: construction & Executed corrected prompt or condition\\
Search repairs a stable-knowledge failure & Generator: internalization & Evidence-free prompt and validated assisted image\\
Failure persists with sufficient valid input & Generator: realization & Compatible rendering, reference, or edit pair\\
Previously mastered capability regresses & Original recipient/replay & Validated prior target\\
Volatile dependency or unresolved diagnosis & Externalize or no-update & Retained access or further probes\\
\bottomrule
\end{tabularx}
\end{table}
\subsection{Which actions remain useful: post-generator refresh}
Post-generator probes cover affected regions, uncertain comparisons, and anchors with the agent, branch inputs, evidence, source images, continuation, evaluator, and seeds fixed. On common tested executable branches $\mathcal A_x^P$, empirical terminal values define a tolerance-aware preferred set:
\begin{equation}
\begin{aligned}
\widehat V_t(x,a)&=\frac{1}{K}\sum_{k=1}^K R_A(\tau_{t,k}^{a},x),\\
\widehat{\mathcal B}_t^\epsilon(x)&=\left\{a\in\mathcal A_x^P:
\widehat V_t(x,a)\geq\max_{b\in\mathcal A_x^P}\widehat V_t(x,b)-\epsilon\right\}.
\end{aligned}
\label{eq:action-boundary}
\end{equation}
Here $\tau_{t,k}^{a}$ forces branch $a$ with the common continuation. Replacing $G_t$ with $\widetilde G_{t+1}$ under matched conditions defines $\widehat{\mathcal B}_{t+1/2}^\epsilon$. Old supervision is \emph{stale} when its action becomes invalid or materially inferior under the updated generator. Preference shifts compare the old and updated preferred sets; staleness evaluates the old supervision action under the updated generator. Rankings depend on differential action improvement.

An internalization-based no-search target additionally requires evidence-free success on reserved new uses of the same knowledge unit, a small remaining Search--Direct gap, and preserved assisted success. Refreshed supervision covers search, query/reference selection, prompt construction, stopping, and supported repair. The agent learns from validated executed trajectories through SFT, then GRPO with the updated generator fixed.

The effect of a generator update on rankings depends on differential gains. Define $\gamma_t(x,a)=\widehat V_{t+1/2}(x,a)-\widehat V_t(x,a)$. Then
\begin{equation}
\begin{split}
\widehat V_{t+1/2}(x,a)-\widehat V_{t+1/2}(x,b)
={}&\widehat V_t(x,a)-\widehat V_t(x,b)\\
&+\gamma_t(x,a)-\gamma_t(x,b).
\end{split}
\label{eq:gap-change}
\end{equation}
Equal gains preserve the pairwise difference; differential gains drive action re-ranking.

\subsection{What the roles learn next: recursive construction and admission}
Writing $z_{t+1/2}$ for post-generator probes, an update attempt is
\begin{equation}
\begin{aligned}
\widetilde G_{t+1}&=\mathcal U_G(G_t;\mathcal C_{A\to G}(z_t,M_t,\Pool)),\\
\widetilde A_{t+1}&=\mathcal U_A(A_t;\mathcal C_{G\to A}(z_{t+1/2},M_t,\Pool),\widetilde G_{t+1}).
\end{aligned}
\label{eq:alternation}
\end{equation}
Here $G\to A$ denotes supervision compiled from generator behavior. Memory stores versioned outcomes for diagnosis, expansion, and replay. After admission, retained roles construct new experience through the fixed collector $\mathcal P_{\Omega_0}$:
\begin{equation}
z_{t+1}=\mathcal P_{\Omega_0}(A_{t+1},G_{t+1},M_{t+1};\Pool)
\longrightarrow\mathcal C_{A\to G}^{t+1}
\longrightarrow\widetilde G_{t+2}.
\label{eq:closure}
\end{equation}
These \emph{construction peers} supply updated evidence, inputs, and outcomes for subsequent curricula. Source controls assess how updated peers improve subsequent curricula while recipient learning and current-partner RL remain active (Section~\ref{sec:source-controls}).

Admission compares target quality and protected generation/understanding capabilities with the retained state on identical held-out subsets, separate from curricula and rewards. Rejected attempts retain the state and consume their budget. Final evaluation compares retained endpoints with initialization and matched controls.

\paragraph{Protected-capability admission.}
For each candidate $\widetilde S_{t+1}$, let $\mathcal D^G_{\mathrm{protect},t}$ and $\mathcal D^A_{\mathrm{protect},t}$ be held-out generation and understanding subsets. Candidate and retained components are evaluated on identical items and settings within the attempt. For protected capability $c$, the guard is
\begin{equation}
\Delta_{c,t}=Q_{c,t}(\widetilde S_{t+1})-Q_{c,t}(S_t)
\geq-\epsilon_c,\qquad \text{for every protected }c.
\label{eq:guard}
\end{equation}
Here $Q_{c,t}$ evaluates the relevant component: generation checks keep raw prompts fixed without agent assistance, and understanding checks invoke the agent's native understanding role without generation. Admission also requires the target-quality criterion. The comparator is the current retained state $S_t$, and rejection leaves that state unchanged.

\FloatBarrier

%% file: sections/13_conceptual_diagnosis.tex
\begin{figure}[!t]
\centering
\includegraphics[width=1\linewidth]{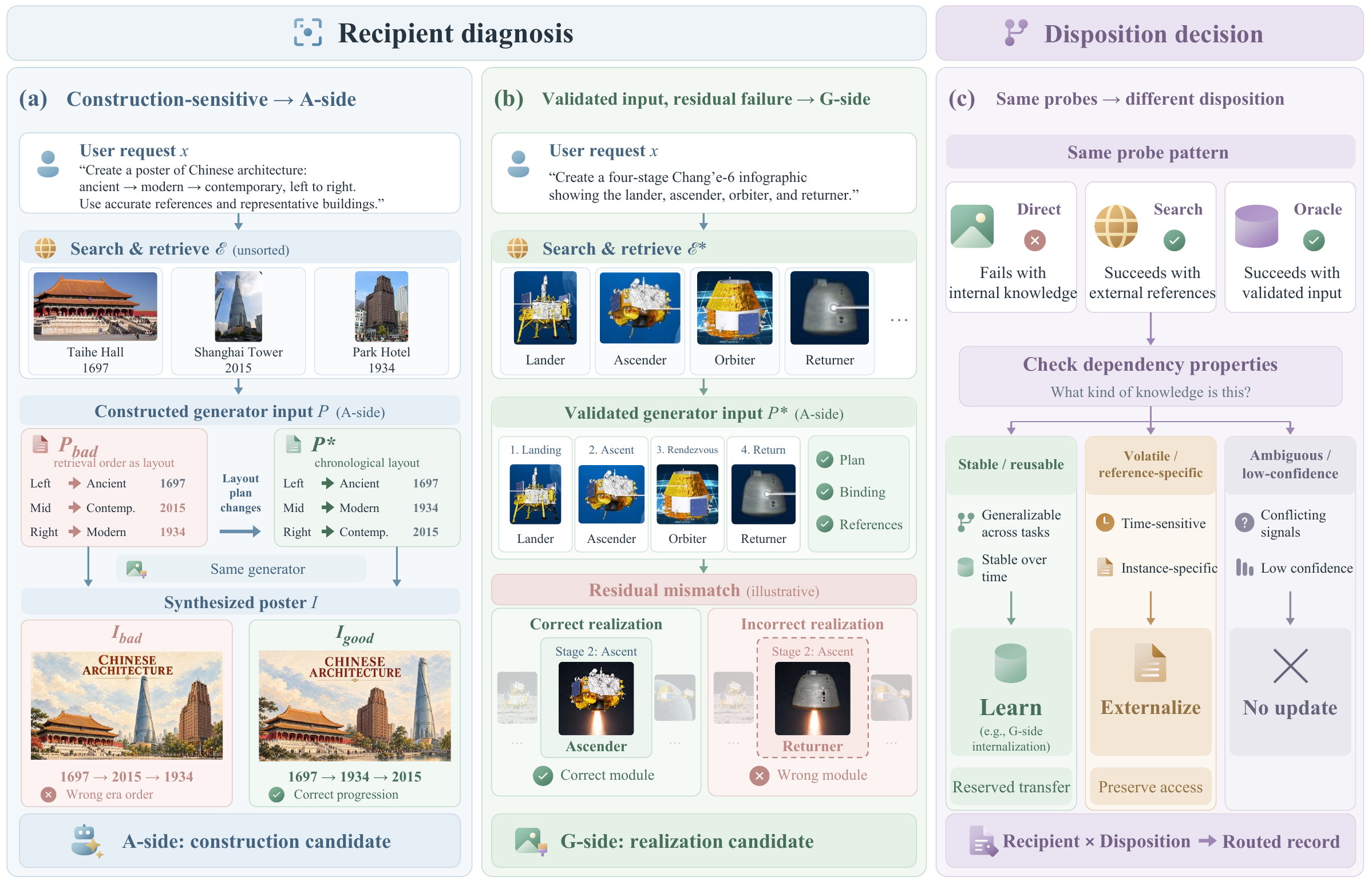}
\caption{Schematic routing decisions. (a) Holding evidence and generator fixed, a corrected layout repairs a chronology error and yields an agent-construction candidate. (b) A module mismatch under checked input yields a generator-realization candidate when a compatible target is available. (c) Learning, externalization, and deferral depend on the dependency's properties and branch outcomes.}
\label{fig:diagnosis}
\end{figure}

%% file: sections/04_experiments.tex
\section{Experimental Setup}
\label{sec:experiments}
\subsection{Backbones, benchmark, and comparison methods}
We evaluate BAGEL, Qwen-Image, and Wan-Image-7B, with mechanism studies on Wan. Qwen-Image uses Qwen2.5-VL as its condition encoder~\citep{qwenimage}; Wan uses full-parameter updates to the trained role. Within each backbone, trained comparisons share initialization, tools, resource allowances, corresponding seeds, and three update attempts. Direct and prompt enhancement provide untrained inference references.

Agentic initialization is the starting tool-using pair. Agent-only adaptation follows Gen-Searcher with a fixed generator and RSIgen's completed-image outcome and trajectory-cost contract; the original Gen-Searcher also uses final-text feedback~\citep{gensearcher}. SearchGen-3 + costs repeats Diffusion-DPO and positive-advantage rejection finetuning for three cycles~\citep{searchgen}. Joint outcome RL trains both roles from terminal outcomes. Uniform reciprocal learning shares RSIgen's role-specific optimization stages and refresh, while changing curriculum allocation. Uniform and failure-locus baselines compare curriculum allocation strategies; one-sided updates evaluate the contribution of each role.

\paragraph{Benchmark and metrics.}
AgentGen-Bench contains 751 requests. We follow its released Overall9 aggregation and \href{https://github.com/HaozheH3/SearchGen}{evaluator}~\citep{searchgen}. Overall9 averages Checklist, Rubric, Prompt Faithfulness, Image Quality, Text Rendering, AI Naturalness, Composition, Physical Plausibility, and Visual Reference. NoSearch (NS) and Search-Intensive (SI) are the benchmark strata; Full aggregates applicable metric scores over each model's scored requests. Published coverage is 738/751 for Nano Banana Pro, 718/751 for GPT-Image-2, and 751/751 for BAGEL and Qwen-Image. Evaluation references are distinct from runtime retrieval inputs and training targets. Dataset and public-result definitions follow the \href{https://huggingface.co/datasets/JasperHaozhe/AgentGen-Bench}{benchmark release} and \href{https://haozheh3.github.io/SearchGen/}{project results}.

\subsection{Matched mechanism studies}
The allocation study uses 4,000 opportunities and matches role-level exposure: 1,600 agent items with 3.2M SFT and 2.4M RL tokens, and 2,000 generator items with 4,000 SFT and 7,200 RL image presentations.

The action-refresh diagnostic uses 200 held-out Test-mini requests, separate from training, tuning, checkpoint selection, and the human-evaluation requests. It compares $(G_0,G_1)$, $(G_1,G_2)$, and $(G_2,G_3)$ under fixed prompts, evidence, action packages, and paired image seeds, with 200 canonical post-states per stage. Action probing uses eight images per action/version. Stale-label and refresh continuation share agent initialization, training requests, the updated rollout generator, and realized token/training-compute budgets. A Stage-2 comparison additionally includes current-generator on-policy learning and matches preparation, rollout, and optimization within a common 100-unit budget: $c=100C/B_{\mathrm{adapt}}$, so one unit is 1\% of the single-stage budget.

Construction-source controls keep recipient learning and current-partner RL active, while freezing the agent, generator, or both as curriculum sources. Every arm uses 1,800 agent and 2,200 generator records per attempt, totaling 5,400 and 6,600. Separate role evaluation uses 200 requests and four image seeds per request/state; agent evaluation fixes $G_0$, and generator evaluation distinguishes direct and assisted inputs.

\subsection{Human evaluation and statistical analysis}
Human evaluation compares RSIgen with SearchGen-3 + costs on the same 200 real-user requests across all three backbones. Three independent blind ratings per pair assess task completion and visual quality, giving 600 judgments per backbone and 1,800 pooled judgments per dimension. Preference credits a tie as half a win.

Main benchmark results aggregate multiple training seeds. The continuation diagnostic uses 10,000 paired request-cluster bootstrap replicates of the evaluated checkpoint outputs, retaining all stages, matched branches, and policy arms for each sampled request. Each replicate recomputes the metrics before taking the equal-weight stage mean; we report percentile 95\% confidence intervals. Resource use is measured by mean search and generation/edit calls per request, including failed calls. Rejected update attempts consume their assigned budgets.

%% file: sections/05_results.tex
\section{Results and Analysis}
\label{sec:results}
\subsection{Benchmark quality and within-backbone comparisons}
\label{sec:effectiveness}
RSIgen reaches AgentGen-Bench Full scores of 39.9, 50.3, and 47.5 on BAGEL, Qwen-Image, and Wan-Image-7B (Table~\ref{tab:main}). Matched training comparisons in Table~\ref{tab:controls} show gains of 3.2, 3.2, and 3.3 points over SearchGen-3 + costs, and 3.7, 3.6, and 3.7 over uniform reciprocal learning. RSIgen also exceeds joint outcome RL on every backbone. Improvements over SearchGen-3 + costs in Table~\ref{tab:controls} are larger on SI (3.4--3.6 points) than NS (0.9--2.2).
\begin{table}[!htbp]
\centering\footnotesize
\caption{AgentGen-Bench overview (0--100). NS/SI: NoSearch/Search-Intensive; Full: Full-slice Overall9. Colors mark \ranklegend{} within each group. Public baseline scores follow the benchmark release; matched training comparisons appear in Table~\ref{tab:controls}.}
\label{tab:main}
\setlength{\tabcolsep}{2.5pt}
\begin{tabular}{lrrrrrrrrr}
\toprule
& \multicolumn{2}{c}{Checklist} & \multicolumn{2}{c}{Visual Ref.} & \multicolumn{2}{c}{Image Quality} & \multicolumn{3}{c}{Overall9}\\
\cmidrule(lr){2-3}\cmidrule(lr){4-5}\cmidrule(lr){6-7}\cmidrule(lr){8-10}
Model & NS & SI & NS & SI & NS & SI & NS & SI & Full\\
\midrule
\rowcolor{TableGroup}
\multicolumn{10}{l}{\emph{Commercial Models}}\\
GPT-Image-2 & \bestcell{84.8} & \bestcell{74.4} & \bestcell{70.9} & \bestcell{70.4} & \bestcell{73.4} & \bestcell{77.4} & \bestcell{78.9} & \bestcell{75.5} & \bestcell{76.0}\\
Grok-Imagine-2.0-Low & \secondcell{83.0} & \secondcell{73.1} & \secondcell{68.7} & \secondcell{68.3} & \secondcell{72.5} & \secondcell{76.9} & \secondcell{76.9} & \secondcell{74.9} & \secondcell{75.2}\\
Qwen-Image-3-Max & 79.1 & 65.8 & 64.0 & 59.6 & 70.3 & 71.1 & 73.7 & 67.6 & 68.4\\
Nano Banana Pro & 78.4 & 60.7 & 63.0 & 55.4 & 70.7 & 68.1 & 73.3 & 63.6 & 65.0\\
Qwen-Image-2 & 76.2 & 49.9 & 60.3 & 42.0 & 70.2 & 62.5 & 71.1 & 54.1 & 56.4\\
\midrule
\rowcolor{TableGroup}
\multicolumn{10}{l}{\emph{Generation Only}}\\
Mage-Flow & \bestcell{64.2} & \secondcell{23.0} & \bestcell{47.1} & \secondcell{16.2} & \bestcell{64.7} & \bestcell{38.7} & \bestcell{61.2} & \bestcell{28.4} & \bestcell{32.8}\\
Flux.2-Klein-9B & \secondcell{63.7} & \bestcell{23.8} & \secondcell{41.1} & \bestcell{16.7} & \secondcell{63.2} & \secondcell{36.1} & \secondcell{58.5} & \secondcell{26.9} & \secondcell{31.1}\\
Flux.2-Klein-4B & 55.8 & 20.1 & 35.3 & 12.7 & 59.7 & 33.0 & 52.3 & 23.9 & 27.7\\
\midrule
\rowcolor{TableGroup}
\multicolumn{10}{l}{\emph{Unified MLLMs}}\\
SenseNova-U1 & 68.1 & 29.3 & 51.1 & 21.0 & 64.3 & 39.5 & 62.8 & 31.6 & 35.7\\
OmniGen2 & 49.1 & 16.4 & 30.6 & 9.5 & 56.4 & 29.2 & 47.4 & 20.4 & 24.0\\
Show-o2 & 32.7 & 8.1 & 19.9 & 3.4 & 47.2 & 30.2 & 34.4 & 17.5 & 19.8\\
BAGEL & 51.5 & 18.7 & 32.1 & 14.2 & 56.2 & 30.4 & 49.1 & 22.2 & 25.8\\
\textbf{BAGEL + RSIgen} & 63.4 & 36.4 & 45.6 & 27.4 & 61.4 & 43.3 & 60.4 & 36.8 & 39.9\\
Qwen-Image & \secondcell{74.9} & 25.0 & \secondcell{60.1} & 17.7 & \secondcell{69.2} & 37.9 & \secondcell{70.0} & 28.6 & 34.1\\
\textbf{Qwen-Image + RSIgen} & \bestcell{77.2} & \bestcell{47.3} & \bestcell{63.4} & \bestcell{37.4} & \bestcell{70.7} & \bestcell{52.3} & \bestcell{73.2} & \bestcell{46.8} & \bestcell{50.3}\\
Wan-Image-7B & 65.3 & 22.3 & 49.3 & 16.3 & 65.1 & 37.2 & 62.3 & 27.4 & 32.1\\
\textbf{Wan-Image-7B + RSIgen} & 70.4 & \secondcell{44.3} & 55.4 & \secondcell{35.4} & 67.2 & \secondcell{50.2} & 67.3 & \secondcell{44.4} & \secondcell{47.5}\\
\bottomrule
\end{tabular}
\end{table}

\begin{table}[!htbp]
\centering\footnotesize
\caption{Matched within-backbone comparisons on AgentGen-Bench. Trained methods share initialization and resource allowances within each backbone; scores aggregate multiple training seeds. NS/SI denote NoSearch/Search-Intensive, and Full is the global Overall9 mean. Direct and prompt enhancement are untrained Wan references; dashes indicate results not reported in this comparison.}
\label{tab:controls}
\setlength{\tabcolsep}{2.4pt}
\begin{tabular}{lrrrrrrrrr}
\toprule
& \multicolumn{3}{c}{BAGEL} & \multicolumn{3}{c}{Qwen-Image} & \multicolumn{3}{c}{Wan-Image-7B}\\
\cmidrule(lr){2-4}\cmidrule(lr){5-7}\cmidrule(lr){8-10}
Method & NS & SI & Full & NS & SI & Full & NS & SI & Full\\
\midrule
Direct & -- & -- & -- & -- & -- & -- & 62.3 & 27.4 & 32.1\\
Prompt enhancement & -- & -- & -- & -- & -- & -- & 64.1 & 31.2 & 35.6\\
Agentic initialization & 54.2 & 25.4 & 29.3 & 70.6 & 35.3 & 40.0 & 64.3 & 33.4 & 37.5\\
A-only / Gen-Searcher adapt. & 56.3 & 29.4 & 33.0 & 71.2 & 39.4 & 43.6 & 65.2 & 36.3 & 40.2\\
SearchGen-3 + costs & 58.2 & 33.4 & 36.7 & 72.3 & 43.2 & 47.1 & 66.2 & 40.8 & 44.2\\
Joint outcome RL & 57.6 & 32.2 & 35.6 & 71.7 & 42.4 & 46.3 & 65.7 & 39.8 & 43.3\\
Uniform reciprocal & 58.6 & 32.8 & 36.2 & 72.1 & 42.8 & 46.7 & 66.3 & 40.3 & 43.8\\
RSIgen & \bestcell{60.4} & \bestcell{36.8} & \bestcell{39.9} & \bestcell{73.2} & \bestcell{46.8} & \bestcell{50.3} & \bestcell{67.3} & \bestcell{44.4} & \bestcell{47.5}\\
\bottomrule
\end{tabular}
\end{table}

Table~\ref{tab:components} breaks down the three backbones and their RSIgen variants by capability. On SI, Checklist improves by 17.7--22.3 points and Text Rendering by 26.2--30.2, compared with 12.9--14.4 for Image Quality. Improvements therefore span task requirements and text rendering as well as visual quality.
\begin{table}[!htbp]
\centering\footnotesize
\caption{Component scores for the three backbones and their RSIgen variants. Ch: Checklist; Ru: Rubric; Pr: Prompt Faithfulness; IQ: Image Quality; Tx: Text Rendering; Nat: AI Naturalness; Co: Composition; Ph: Physical Plausibility; VR: Visual Reference. Overall9 appears in Table~\ref{tab:main}.}
\label{tab:components}
\setlength{\tabcolsep}{2.7pt}
\begin{tabular}{lrrrrrrrrr}
\toprule
Model & Ch & Ru & Pr & IQ & Tx & Nat & Co & Ph & VR\\
\midrule
\rowcolor{TableGroup}
\multicolumn{10}{l}{\textbf{A. NoSearch (NS)}}\\
\midrule
BAGEL & 51.5 & 49.2 & 37.7 & 56.2 & 16.1 & 48.8 & 62.8 & 58.5 & 32.1\\
\textbf{BAGEL + RSIgen} & 63.4 & 62.1 & 58.3 & 61.4 & 50.6 & 60.2 & 68.4 & 73.2 & 45.6\\
Qwen-Image & 74.9 & 72.0 & 65.5 & 69.2 & 61.1 & 61.7 & 77.0 & 78.6 & 60.1\\
\textbf{Qwen-Image + RSIgen} & 77.2 & 76.1 & 72.4 & 70.7 & 68.4 & 69.2 & 80.1 & 81.3 & 63.4\\
Wan-Image-7B & 65.3 & 63.2 & 57.4 & 65.1 & 54.3 & 57.2 & 73.4 & 75.1 & 49.3\\
\textbf{Wan-Image-7B + RSIgen} & 70.4 & 69.2 & 65.3 & 67.2 & 60.4 & 63.2 & 76.1 & 75.4 & 55.4\\
\midrule
\rowcolor{TableGroup}
\multicolumn{10}{l}{\textbf{B. Search-Intensive (SI)}}\\
\midrule
BAGEL & 18.7 & 17.7 & 12.4 & 30.4 & 2.2 & 25.0 & 32.4 & 38.1 & 14.2\\
\textbf{BAGEL + RSIgen} & 36.4 & 35.2 & 31.4 & 43.3 & 29.5 & 37.2 & 44.3 & 44.8 & 27.4\\
Qwen-Image & 25.0 & 23.8 & 17.4 & 37.9 & 9.2 & 29.9 & 41.0 & 45.5 & 17.7\\
\textbf{Qwen-Image + RSIgen} & 47.3 & 46.2 & 42.4 & 52.3 & 39.4 & 45.2 & 55.1 & 55.7 & 37.4\\
Wan-Image-7B & 22.3 & 21.2 & 15.4 & 37.2 & 10.3 & 30.2 & 41.3 & 51.2 & 16.3\\
\textbf{Wan-Image-7B + RSIgen} & 44.3 & 43.2 & 39.4 & 50.2 & 36.5 & 43.2 & 53.4 & 53.2 & 35.4\\
\bottomrule
\end{tabular}
\end{table}

\subsection{Human evaluation on real-user requests}
\label{sec:human-evaluation}
RSIgen receives more wins than losses for both dimensions on every backbone (Table~\ref{tab:human-evaluation-full}). Tie-adjusted task-completion preference ranges from 68.5\% to 71.4\%, and visual-quality preference from 63.5\% to 66.1\%. The pooled scores are 70.0\% and 64.9\%, respectively.

\begin{table}[!htbp]
\centering\small
\setlength{\tabcolsep}{4pt}
\caption{Blind human evaluation against SearchGen-3 + costs on the same 200 real-user requests. Each backbone has 600 judgments per dimension (three ratings per pair); Overall pools 1,800 judgments on those same requests. W/T/L is from RSIgen's perspective; preference is $100(W+0.5T)/(W+T+L)$.}
\label{tab:human-evaluation-full}
\begin{tabular}{lcrcr}
\toprule
& \multicolumn{2}{c}{Task completion} & \multicolumn{2}{c}{Visual quality}\\
\cmidrule(lr){2-3}\cmidrule(lr){4-5}
Backbone & W / T / L & Pref. & W / T / L & Pref.\\
\midrule
BAGEL & 312 / 198 / 90 & 68.5\% & 264 / 234 / 102 & 63.5\%\\
Qwen-Image & 324 / 194 / 82 & 70.2\% & 276 / 228 / 96 & 65.0\%\\
Wan-Image-7B & 336 / 185 / 79 & 71.4\% & 288 / 217 / 95 & 66.1\%\\
\midrule
\rowcolor{TableGroup}
\textbf{Overall} & 972 / 577 / 251 & \textbf{70.0\%} & 828 / 679 / 293 & \textbf{64.9\%}\\
\bottomrule
\end{tabular}
\end{table}
\subsection{Diagnostic allocation and quality--tool-use tradeoffs}
\label{sec:mechanism}
Table~\ref{tab:ablation} compares curriculum strategies and role updates on Wan. Agent-only and generator-only updates reach 40.2 and 41.7, compared with RSIgen's 47.5. Uniform reciprocal and failure-locus-only routing reach 43.8 and 44.1. Removing post-generator re-probing lowers Full to 45.0; removing capability memory or pool expansion lowers it to 45.9 or 45.3.
\begin{table}[!htbp]
\centering\small
\caption{Wan ablations on AgentGen-Bench. Full measures quality; S and G/E count inference searches and generations/edits per request.}
\label{tab:ablation}
\setlength{\tabcolsep}{3.5pt}
\begin{tabular}{lrrr}
\toprule
Variant & Full $\uparrow$ & S & G/E\\
\midrule
Full RSIgen & \secondcell{47.5} & \bestcell{1.16} & \bestcell{1.30}\\
A-only updates & 40.2 & 1.59 & 1.45\\
G-only updates & 41.7 & 1.76 & 1.51\\
Uniform reciprocal & 43.8 & 1.41 & 1.42\\
Failure-locus-only routing & 44.1 & 1.51 & 1.44\\
No post-G re-probe & 45.0 & 1.53 & 1.39\\
No capability memory & 45.9 & 1.31 & 1.36\\
No full-pool expansion & 45.3 & \secondcell{1.29} & \secondcell{1.35}\\
No trajectory costs & \bestcell{48.0} & 1.86 & 1.56\\
\bottomrule
\end{tabular}
\end{table}

\paragraph{Allocation under matched role exposure.}
The allocation study holds agent and generator training volumes fixed (Section~\ref{sec:experiments}). In Table~\ref{tab:measured-drr}, DRR reaches Full 47.4, gaining 1.4 over compatible-uniform (46.0) and 1.1 over failure-locus (46.3). Searches and generations/edits decrease from 1.29/1.37 to 1.21/1.32 relative to compatible-uniform. Thus the gain persists at matched role-level item, token, and image exposure.
\begin{table}[!htbp]
\centering\small
\setlength{\tabcolsep}{3pt}
\caption{Wan allocation comparison with matched role-level training exposure. $\Delta$ is relative to compatible-uniform.}
\label{tab:measured-drr}
\begin{tabular}{lrrrr}
\toprule
Strategy & Full $\uparrow$ & $\Delta$ Full & Searches $\downarrow$ & Gen./edits $\downarrow$\\
\midrule
Compatible-uniform & 46.0 & 0.0 & 1.29 & 1.37\\
Failure-locus & \secondcell{46.3} & \secondcell{+0.3} & \secondcell{1.27} & \secondcell{1.36}\\
DRR & \bestcell{47.4} & \bestcell{+1.4} & \bestcell{1.21} & \bestcell{1.32}\\
\bottomrule
\end{tabular}
\par\vspace{3pt}
\begin{minipage}{0.97\linewidth}\footnotesize
All arms use 4,000 opportunities. Agent: 1,600 training entries, 3.20M SFT tokens and 2.40M RL tokens. Generator: 2,000 entries, 4,000 SFT and 7,200 RL image presentations. Calls are per request.
\end{minipage}
\end{table}
\paragraph{Inference quality and tool use.}
The no-cost variant reaches Full 48.0 versus RSIgen's 47.5, with searches rising from 1.16 to 1.86 and generations/edits from 1.30 to 1.56 per request: 60.3\% and 20.0\% increases from the displayed averages. Trajectory costs trade 0.5 quality points for fewer inference calls. RSIgen also achieves higher quality with slightly fewer calls than SearchGen-3 + costs across all three backbones (Table~\ref{tab:resource}).
\begin{table}[!htbp]
\centering\small
\caption{Inference-call averages for SearchGen-3 + costs and RSIgen. Values are mean search and generation/edit calls per request.}
\label{tab:resource}
\begin{tabular}{llrrr}
\toprule
Backbone & Method & Full & Searches & Generations/edits\\
\midrule
BAGEL & SearchGen-3 + costs & \secondcell{36.7} & \secondcell{1.21} & \secondcell{1.32}\\
BAGEL & RSIgen & \bestcell{39.9} & \bestcell{1.18} & \bestcell{1.31}\\
Qwen-Image & SearchGen-3 + costs & \secondcell{47.1} & \secondcell{1.21} & \secondcell{1.31}\\
Qwen-Image & RSIgen & \bestcell{50.3} & \bestcell{1.13} & \bestcell{1.28}\\
Wan-Image-7B & SearchGen-3 + costs & \secondcell{44.2} & \secondcell{1.26} & \secondcell{1.33}\\
Wan-Image-7B & RSIgen & \bestcell{47.5} & \bestcell{1.16} & \bestcell{1.30}\\
\bottomrule
\end{tabular}
\end{table}

\subsection{Action utility shifts and post-generator adaptation}
\label{sec:action-refresh}
\paragraph{Matched branches and evaluated populations.}
The diagnostic distinguishes changes in action utility from the policy's response to new supervision. Old/new generator comparisons fix the agent, enhanced prompts, search evidence, references, source images, continuation, evaluator, and paired seeds. Search branches retain their artifacts and logical call costs, including when retrieval yields no usable evidence. A canonical post-state uses the same source image for both checkpoints and continuation arms. Accept returns that image; regeneration and reference-conditioned regeneration use the fixed input packages; LocalEdit evaluates identifiable localized repairs supported by both checkpoints.

There are 200 evaluated pre-generation requests and 200 post-states per stage. Across 600 stage--request observations, text and image search yield no usable evidence in 162 (27.0\%) and 204 (34.0\%) cases, with zero infrastructure failures after retries. LocalEdit is eligible for 126, 121, and 116 states across the stages. Empty retrieval outcomes remain included in action values. End-to-end Mini O9 evaluates the separate complete policy workflow, including its chosen queries and induced state distribution.

\paragraph{Generator learning changes relative action utility.}
Table~\ref{tab:action-refresh}A reports preference shifts in 22.2\% of pre-generation and 20.8\% of post-generation decisions; old labels are stale in 17.0\% and 15.7\%. A shift changes the preferred action set, whereas staleness makes the old supervision action invalid or materially inferior. Pre-generation shift counts are 49, 44, and 40, and stale counts are 38, 34, and 30; post-generation shift counts are 45, 42, and 38, and stale counts are 34, 31, and 29. Headline rates average the stage proportions.

Search-to-Direct accounts for $69/133=51.9\%$ of pre-generation shifts; Regenerate-to-LocalEdit accounts for $49/125=39.2\%$ of post-generation shifts (Table~\ref{tab:action-transitions}). Accept's image quality is checkpoint-invariant, so shifts involving Accept reflect changes in the relative benefit of intervention.
\begin{table}[!htbp]
\centering\footnotesize
\caption{Wan action refresh on Test-mini. A: branch changes over 200 requests and 200 canonical post-states per stage, with 95\% intervals in brackets. B: matched continuation with the updated generator; the final column gives refresh-minus-stale 95\% intervals. All estimates average three stages. Intervals use 10,000 paired request-cluster bootstrap replicates.}
\label{tab:action-refresh}
\setlength{\tabcolsep}{4pt}
\begin{tabular*}{\linewidth}{@{\extracolsep{\fill}}lrrr@{}}
\toprule
\rowcolor{TableGroup}
\multicolumn{4}{l}{\textbf{A. Action changes after generator updates}}\\
Decision & Evaluated per stage & Preference shift (\%) & Stale labels (\%)\\
\midrule
Pre-generation & 200 & 22.2 [18.1, 26.5] & 17.0 [13.4, 20.9]\\
Post-generation & 200 & 20.8 [16.9, 25.1] & 15.7 [12.3, 19.3]\\
\bottomrule
\end{tabular*}
\vspace{5pt}
\begin{tabular*}{\linewidth}{@{\extracolsep{\fill}}lrrrr@{}}
\toprule
\rowcolor{TableGroup}
\multicolumn{5}{l}{\textbf{B. Policy adaptation and paired differences}}\\
Metric & Old $A_j$ & Stale labels & Refresh & $\Delta$: 95\% CI\\
\midrule
Pre regret $\downarrow$ & 7.63 & \secondcell{5.37} & \bestcell{3.30} & [$-2.60$, $-1.60$]\\
Post regret $\downarrow$ & 6.10 & \secondcell{4.60} & \bestcell{3.00} & [$-2.00$, $-1.20$]\\
Mini O9 $\uparrow$ & 44.77 & \secondcell{47.37} & \bestcell{50.20} & [1.80, 3.90]\\
Pre Track (\%) $\uparrow$ & 34.0 & \secondcell{52.9} & \bestcell{78.2} & [18.00, 32.00]\\
Post Agree (\%) $\uparrow$ & 67.3 & \secondcell{75.1} & \bestcell{83.4} & [5.00, 12.00]\\
LocalEdit precision (\%) $\uparrow$ & 62.6 & \secondcell{72.7} & \bestcell{83.8} & [7.00, 16.00]\\
LocalEdit recall (\%) $\uparrow$ & 49.5 & \secondcell{61.4} & \bestcell{76.3} & [10.00, 20.00]\\
\bottomrule
\end{tabular*}
\par\vspace{3pt}
\begin{minipage}{\linewidth}\footnotesize
Regret uses reward units; percentage-metric differences use percentage points. Pre Track conditions on pre-generation shifts, Post Agree uses all post-states, and LocalEdit precision/recall use eligible states.
\end{minipage}
\end{table}
\begin{table}[!htbp]
\centering\small
\setlength{\tabcolsep}{3pt}
\caption{Grouped action transitions pooled across stages. Shares divide by 133 pre-generation or 125 post-generation shifts. SearchModeSwitch includes switches involving HybridSearch; Repair groups admissible repair actions.}
\label{tab:action-transitions}
\begin{tabular}{llrr}
\toprule
Phase & Transition & Count & Share (\%)\\
\midrule
Pre & Search$\to$Direct & 69 & 51.9\\
Pre & Direct$\to$Search & 15 & 11.3\\
Pre & SearchModeSwitch & 24 & 18.0\\
Pre & Search$\to$PromptEnhance & 13 & 9.8\\
Pre & PromptEnhance$\to$Search & 12 & 9.0\\
Post & Regenerate$\to$LocalEdit & 49 & 39.2\\
Post & ReferenceRegenerate$\to$LocalEdit & 21 & 16.8\\
Post & LocalEdit$\to$Regenerate & 16 & 12.8\\
Post & Accept$\to$Repair & 18 & 14.4\\
Post & Repair$\to$Accept & 21 & 16.8\\
\bottomrule
\end{tabular}
\end{table}
\paragraph{Measured branch examples.}
Two branch-level examples illustrate differential action gains. In one, Direct quality rises from 0.38 to 0.75 while Search remains at 0.82--0.83, narrowing the Search--Direct gap from 0.44 to 0.08. In a repair case, LocalEdit's quality-minus-incremental-cost value rises from 0.54 to 0.77 and overtakes regeneration (0.62 to 0.64); Accept remains at 0.52 and reference-conditioned regeneration changes from 0.59 to 0.64.

\paragraph{Refreshed supervision improves continuation.}
Stale-label and refresh arms share initialization, requests, the updated rollout generator, and realized training compute. Refresh reduces pre/post regret from 5.37/4.60 to 3.30/3.00 and increases three-stage Mini O9 from 47.37 to 50.20. Table~\ref{tab:action-refresh}B also reports complementary alignment measures: Pre Track conditions on pre-generation shift cases, Post Agree uses all evaluated post-states, and LocalEdit precision/recall condition on eligible states. The direction is consistent across all three stages: refresh improves Mini O9 over stale continuation by 2.8, 2.9, and 2.8 points, and Pre Track by 25.9, 25.8, and 24.0 percentage points, respectively.

Regret averages the best tested action value minus the selected action value in the reward units of Eq.~\eqref{eq:rewarda}. Pre Track and Post Agree measure preference membership under the updated generator. Table~\ref{tab:action-refresh} reports paired request-cluster 95\% intervals, with [1.8, 3.9] for the Mini O9 gain.

\paragraph{Comparison with current-generator on-policy learning.}
The Stage-2 comparison charges preparation as well as rollout and optimization within the common 100-unit budget (Table~\ref{tab:measured-refresh}). Refresh reaches Mini O9 50.0, versus on-policy learning's 49.1 and stale supervision's 47.2. Pre/post regret decreases from 4.0/3.7 to 3.5/3.2 relative to on-policy learning.
\begin{table}[!htbp]
\centering
\sbox{\rsigenRefreshTableBox}{%
\begin{minipage}{0.48\linewidth}
\centering
\renewcommand{\arraystretch}{1.25}
\centering\footnotesize
\setlength{\tabcolsep}{2pt}
\begin{tabular}{@{}lrrr@{}}
\toprule
Metric & Stale & \shortstack{Current-G\\on-policy} & Refresh\\
\midrule
\rowcolor{TableGroup}
\multicolumn{4}{l}{\emph{Quality and action selection}}\\
Mini O9 $\uparrow$ & 47.2 & \secondcell{49.1} & \bestcell{50.0}\\
$\Delta$ O9 & -1.9 & \secondcell{0.0} & \bestcell{+0.9}\\
Pre regret & 5.40 & \secondcell{4.00} & \bestcell{3.50}\\
Post regret & 4.60 & \secondcell{3.70} & \bestcell{3.20}\\
\midrule
\rowcolor{TableGroup}
\multicolumn{4}{l}{\emph{Adaptation cost}}\\
Preparation & 5 & 0 & 20\\
Rollout & 65 & 74 & 55\\
Optimization & 30 & 26 & 25\\
Total & 100 & 100 & 100\\
\bottomrule
\end{tabular}
\end{minipage}}
\setlength{\rsigenPanelBodyHeight}{\dimexpr\ht\rsigenRefreshTableBox+\dp\rsigenRefreshTableBox\relax}
\sbox{\rsigenRecursionCaptionBox}{%
\begin{minipage}[t]{0.48\linewidth}
\captionsetup{type=figure,position=top,skip=4pt}
\caption{Wan curriculum-source controls on AgentGen-Bench Full. Recipients keep learning with current-partner RL while the named construction sources are frozen.}
\label{fig:recursion}
\end{minipage}}
\sbox{\rsigenRefreshCaptionBox}{%
\begin{minipage}[t]{0.48\linewidth}
\captionsetup{type=table,position=top,skip=4pt}
\caption{Wan Stage-2 adaptation at a matched total budget of 100, including preparation, rollout, and optimization. On-policy uses the current generator; $\Delta$ O9 is relative to this arm. One cost unit is 1\% of the common single-stage budget.}
\label{tab:measured-refresh}
\end{minipage}}
\setlength{\rsigenPanelCaptionHeight}{\dimexpr\ht\rsigenRecursionCaptionBox+\dp\rsigenRecursionCaptionBox\relax}
\ifdim\dimexpr\ht\rsigenRefreshCaptionBox+\dp\rsigenRefreshCaptionBox\relax>\rsigenPanelCaptionHeight
  \setlength{\rsigenPanelCaptionHeight}{\dimexpr\ht\rsigenRefreshCaptionBox+\dp\rsigenRefreshCaptionBox\relax}
\fi
\begin{minipage}[t]{0.48\linewidth}
\vspace{0pt}
\centering
\begin{minipage}[t][\rsigenPanelCaptionHeight][t]{\linewidth}
\usebox{\rsigenRecursionCaptionBox}
\end{minipage}\par\nointerlineskip
\includegraphics[height=\rsigenPanelBodyHeight,width=\linewidth,keepaspectratio]{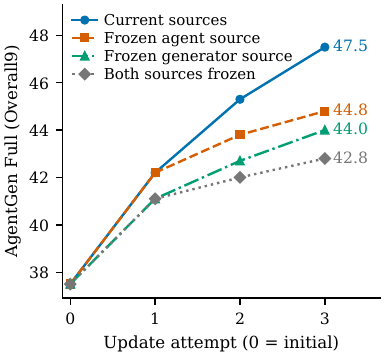}
\end{minipage}\hfill
\begin{minipage}[t]{0.48\linewidth}
\vspace{0pt}
\centering
\begin{minipage}[t][\rsigenPanelCaptionHeight][t]{\linewidth}
\usebox{\rsigenRefreshCaptionBox}
\end{minipage}\par\nointerlineskip
\usebox{\rsigenRefreshTableBox}
\end{minipage}
\end{table}

\subsection{Updated construction sources and recursive curricula}
\label{sec:source-controls}

\paragraph{Controls retain recipient learning.}
Freeze-A uses $A_0$ to construct curriculum evidence and inputs; Freeze-G uses $G_0$ for curriculum-side outcomes, including post-generator diagnosis. Freeze-both fixes both sources, while the full arm uses current sources. Trainable recipients continue learning in every arm, and agent RL uses its current updated generator. Source opportunities, retrieval snapshots, compiler, optimization stages, and resource caps define the matched comparison. Each arm derives its memory, targets, and scores from its designated sources. With common randomness, current and Freeze-A share the first attempt because both initially use $A_0$.

Current sources reach 47.5 on Wan, compared with 42.8 under dual freezing. Freezing only the agent gives 44.8; freezing only the generator gives 44.0. Current sources improve Full by 4.7, 3.1, and 2.2 points across the three attempts (Figure~\ref{fig:recursion}).

\paragraph{Usable curriculum yield at fixed training sizes.}
Each source arm considers 2,000 requests per phase, yielding 4,000 generator-target candidates before the generator update and 3,000 agent-target candidates afterward. Validity is the valid-to-candidate ratio. Cross-attempt construction totals count target entries, separately from selected training records and repeated optimization exposure. New, refreshed, and existing-use records partition the selected curriculum; existing use includes replay and first use of previously available targets.

Across three attempts, current sources yield 5,580/7,480 valid agent/generator targets, compared with 4,430/6,740 under dual freezing, with 1,950 versus zero refreshed agent targets. Every arm trains on 5,400/6,600 agent/generator items (Table~\ref{tab:measured-source-totals}). Quality improves alongside validity and freshness at fixed selected training sizes.
\begin{table}[!htbp]
\centering\footnotesize
\setlength{\tabcolsep}{3pt}
\caption{Wan curriculum construction and supervision changes across the four source controls. Counts sum target entries across three attempts. Valid and new-valid counts describe construction; refreshed counts describe updated targets actually used for training.}
\label{tab:measured-source-totals}
\begin{tabular}{lrrrrrr}
\toprule
& \multicolumn{3}{c}{Agent curricula} & \multicolumn{3}{c}{Generator curricula}\\
\cmidrule(lr){2-4}\cmidrule(lr){5-7}
Source & Valid & New valid & Refreshed & Valid & New valid & Refreshed\\
\midrule
Current & 5,580 & 2,660 & 1,950 & 7,480 & 4,140 & 970\\
Freeze-A & 5,280 & 2,560 & 1,660 & 7,060 & 3,820 & 790\\
Freeze-G & 4,690 & 2,270 & 400 & 6,960 & 3,680 & 680\\
Freeze-both & 4,430 & 2,120 & 0 & 6,740 & 3,540 & 570\\
\bottomrule
\end{tabular}
\par\vspace{3pt}
\begin{minipage}{0.97\linewidth}\footnotesize
Each arm uses 5,400 agent and 6,600 generator training entries over the three attempts.
\end{minipage}
\end{table}
\subsection{Role trajectories and curriculum evolution}
\label{sec:role-curriculum}

\begin{figure}[!htbp]
\centering
\includegraphics[width=0.92\linewidth]{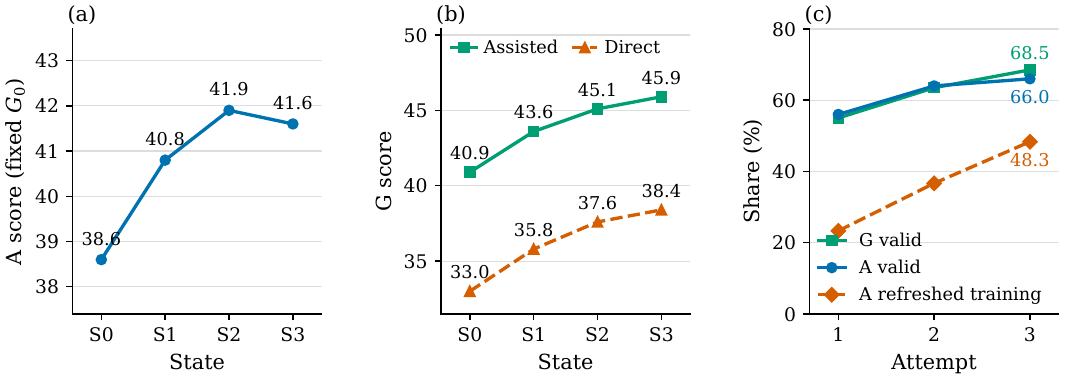}
\caption{Wan role performance and curriculum validity across retained states. Evaluation uses 200 requests and four image seeds per request/state; agent evaluation fixes $G_0$. Role scores and curriculum units are defined in Section~\ref{sec:role-curriculum}.}
\label{fig:measured-evolution}
\end{figure}
\paragraph{Performance under specified partners and inputs.}
Figure~\ref{fig:measured-evolution} tracks role quality and curriculum validity across four retained states on 200 requests with four image seeds per request/state. Agent evaluation fixes $G_0$; generator evaluation uses direct and evidence-assisted inputs. Agent quality starts at 38.6 and reaches 41.9 at $S_2$ and 41.6 at $S_3$. Assisted/direct generator quality rises from 40.9/33.0 to 45.9/38.4.

\paragraph{Validity, freshness, and skill composition.}

\begin{figure}[!htbp]
\centering
\includegraphics[width=0.92\linewidth]{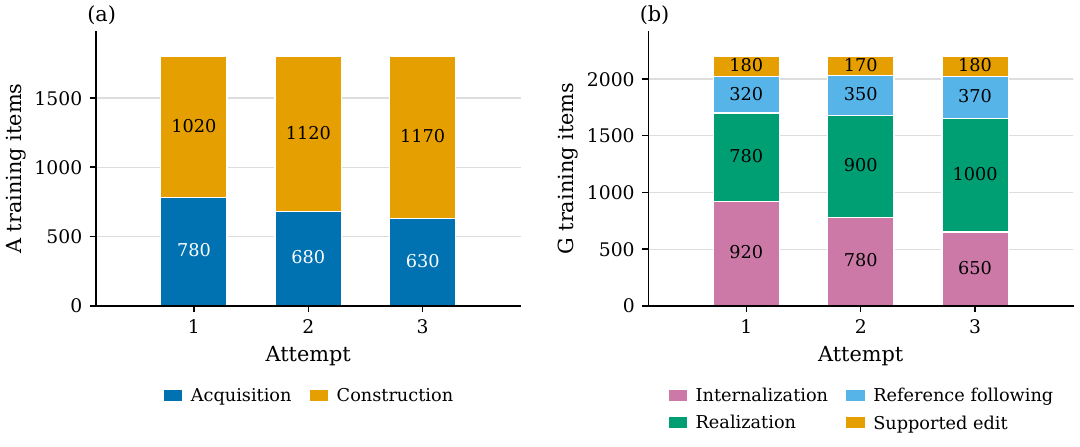}
\caption{Current-source training curricula across three Wan update attempts. Each bar counts selected items by recipient subtype, with totals of 1,800 agent and 2,200 generator items per attempt. Counts are selected training records, including replay.}
\label{fig:measured-curriculum}
\end{figure}
For current sources, agent/generator construction validity rises from 56.0\%/55.0\% to 66.0\%/68.5\%. Refreshed agent supervision grows from $420/1800=23.3\%$ to $870/1800=48.3\%$. The first and last attempts yield 1,500 and 1,180 newly valid generator targets, and 1,080 and 650 newly valid agent targets, respectively.

Figure~\ref{fig:measured-curriculum} shows the selected records by skill. Agent construction grows from 1,020 to 1,170 while acquisition falls from 780 to 630; generator realization grows from 780 to 1,000 while internalization falls from 920 to 650. Each attempt uses 1,800 agent and 2,200 generator entries.

\subsection{Qualitative results}
\label{sec:qualitative}
\input{figures/measured_visual_layout}

\begin{figure}[p]
\centering
\includegraphics[width=\linewidth,height=0.90\textheight,keepaspectratio]{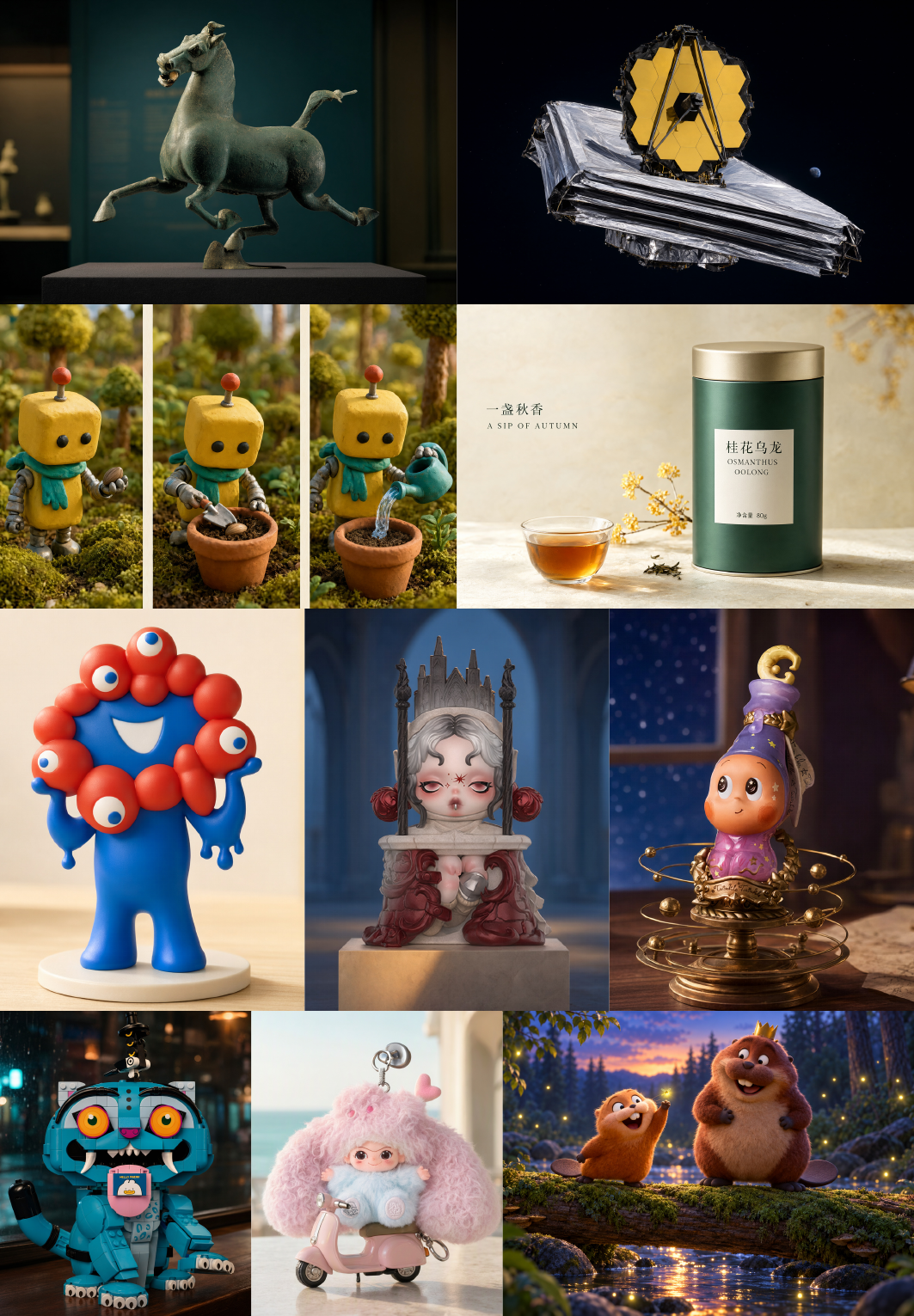}
\caption{Additional RSIgen visual examples, including object depiction, sequential character scenes, product presentation, and stylized character designs.}
\label{fig:visual-showcase}
\end{figure}
Figure~\ref{fig:measured-visual} compares five requests with their complete evaluation prompts and 25 model outputs: calligraphy, TLS 1.3, flags, astronomy, and coordinated design variations. RSIgen separates client/server columns and labels protocol stages in the TLS example, and renders four orrery designs in a consistent grid. The other requests illustrate text, entity appearance, and scene-composition requirements. These examples complement the aggregate benchmark and human judgments.

Figure~\ref{fig:visual-showcase} shows additional visual examples, including object depiction, sequential character scenes, product presentation, and stylized characters.


%% file: figures/measured_visual_layout.tex
\begin{figure}[!htbp]
\centering
\includegraphics[width=\linewidth]{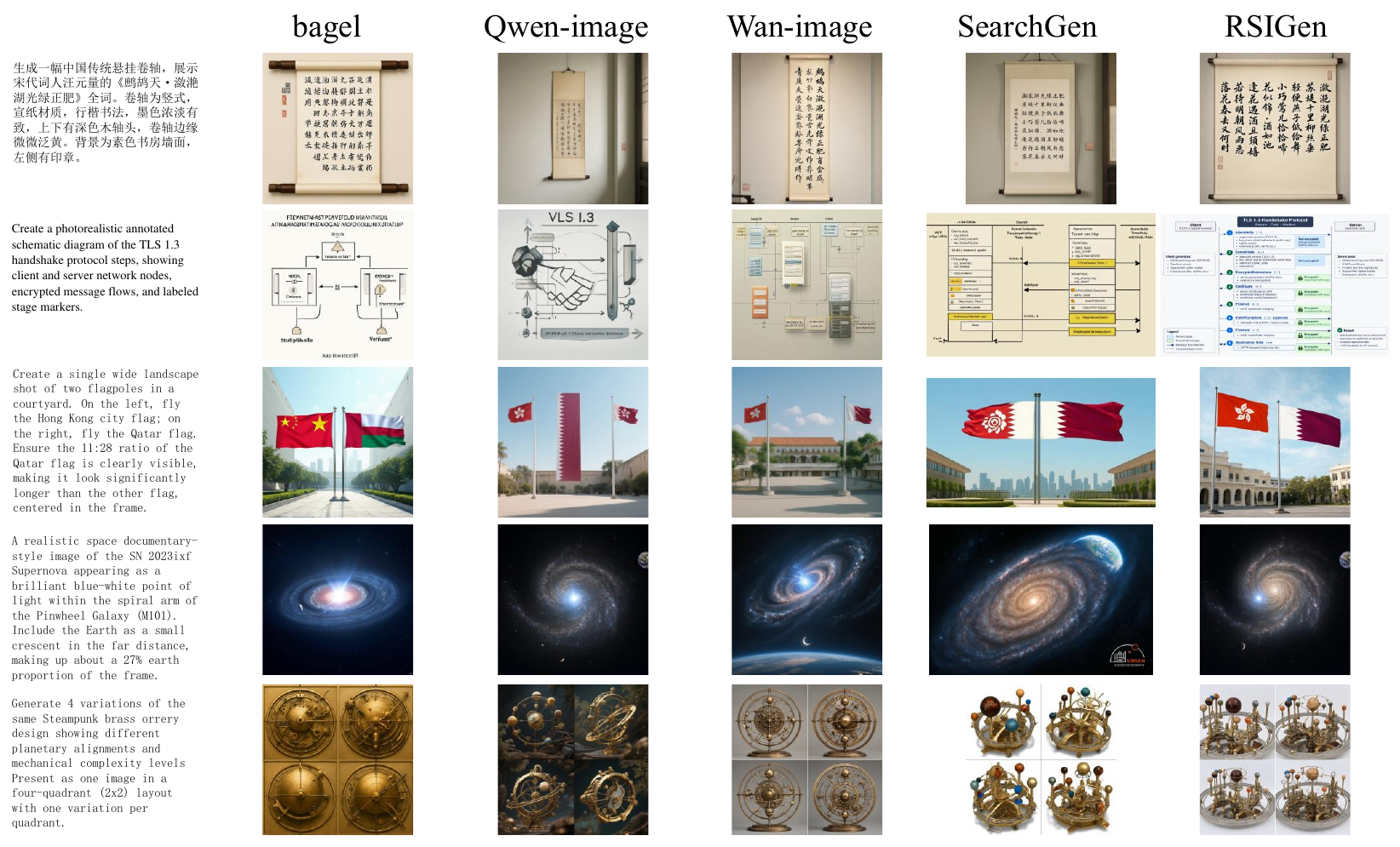}
\caption{Five-request comparison with complete prompts at left and BAGEL, Qwen-Image, Wan-Image, SearchGen, and RSIgen outputs in columns. Table~\ref{tab:controls} reports aggregate comparisons.}
\label{fig:measured-visual}
\end{figure}

%% file: sections/06_discussion.tex
\section{Limitations}
\label{sec:limitations}
Current agentic visual generation systems, whether built from separate understanding and generation models or a unified model, often understand task requirements more reliably than they can realize them in generated images. RSIgen uses this capability asymmetry to improve collaboration between understanding and generation, translating stronger understanding into more effective supervision and better system-level performance. Nevertheless, the extent of further improvement remains constrained by the generative capabilities of the underlying models. Larger advances therefore also depend on improving these models' ability to translate understood requirements into faithful visual outputs.

%% file: sections/07_conclusion.tex
\section{Conclusion}
RSIgen connects diagnosis-guided learning allocation, supervision refresh, and reciprocal curriculum construction. It improves AgentGen-Bench quality across three backbones, with complementary human preferences on 200 real-user requests. Wan controls show allocation and refresh gains at matched exposure and adaptation budget, while source controls connect recursive improvement to target validity and supervision freshness at fixed curriculum sizes. The role trajectories highlight the importance of adapting coordination as capabilities change.